\documentclass[letterpaper, 10 pt, conference]{ieeeconf}  

\IEEEoverridecommandlockouts                              

\makeatletter
\let\NAT@parse\undefined
\makeatother

\usepackage{amsmath,amsfonts}
\usepackage{algorithmic}
\usepackage{array}
\usepackage{textcomp}
\usepackage{stfloats}
\usepackage{url}
\usepackage{verbatim}
\usepackage{graphicx}
\usepackage{balance}

\usepackage[numbers,sectionbib,sort&compress]{natbib}
\usepackage{cite}
\usepackage[printonlyused,withpage,nolist,nohyperlinks]{acronym}
\begin{acronym}

\acro{2D}{two-dimensional}

\acro{3D}{three-dimensional}

\acro{AHRS}{attitude and heading reference system}

\acro{AUV}{autonomous underwater vehicle}
\acro{AGV}{autonomous ground vehicle}

\acro{CPP}{Chinese Postman Problem}

\acro{DoF}{degree of freedom}
\acrodefplural{DoF}[DoFs]{degrees of freedom}

\acro{DVL}{Doppler velocity log}

\acro{FSM}{finite state machine}

\acro{IMU}{inertial measurement unit}

\acro{LBL}{Long Baseline}

\acro{MCM}{mine countermeasures}

\acro{MDP}{Markov decision process}
\acrodefplural{MDP}[MDPs]{Markov decision processes}

\acro{POMDP}{Partially Observable Markov Decision Process}
\acrodefplural{POMDP}[POMDPs]{Partially Observable Markov Decision Processes}

\acro{PRM}{Probabilistic Roadmap}
\acrodefplural{PRM}[PRM]{Probabilistic Roadmaps}

\acro{ROI}{region of interest}
\acrodefplural{ROI}[ROIs]{regions of interest}

\acro{ROS}{Robot Operating System}

\acro{ROV}{remotely operated vehicle}

\acro{RRT}{Rapidly-exploring Random Tree}
\acrodefplural{RRT}[RRTs]{Rapidly-exploring Random Trees}

\acro{SLAM}{Simultaneous Localisation and Mapping}

\acro{SSE}{sum of squared errors}

\acro{STOMP}{Stochastic Trajectory Optimization for Motion Planning}

\acro{TRN}{Terrain-Relative Navigation}

\acro{UAV}{unmanned aerial vehicles}

\acro{USBL}{Ultra-Short Baseline}

\acro{IPP}{informative path planning}

\acro{FoV}{field of view}
\acrodefplural{FoV}[FoVs]{fields of view}

\acro{CDF}{cumulative distribution function}

\acro{ML}{maximum likelihood}

\acro{RMSE}{Root Mean Squared Error}
\acro{MLL}{Mean Log Loss}

\acro{GP}{Gaussian Process}
\acrodefplural{GP}[GPs]{Gaussian Processes}

\acro{KF}{Kalman Filter}

\acro{IP}{Interior Point}
\acro{BO}{Bayesian Optimization}

\acro{SE}{squared exponential}

\acro{UI}{uncertain input}

\acro{MCL}{Monte Carlo Localisation}
\acro{AMCL}{Adaptive Monte Carlo Localisation}

\acro{SSIM}{Structural Similarity Index}

\acro{MAE}{Mean Absolute Error}
\acro{RMSE}{Root Mean Squared Error}
\acro{AUSE}{Area Under the Sparsification Error curve}

\acro{AL}{active learning}
\acro{DL}{deep learning}
\acro{CNN}{convolutional neural network}

\acro{MC}{Monte-Carlo}

\acro{GSD}{ground sample distance}

\acro{BALD}{Bayesian active learning by disagreement}

\acro{fCNN}{fully convolutional neural network}
\acro{FCN}{fully convolutional neural network}

\acro{CMA-ES}{covariance matrix adaptation evolution strategy}

\acro{FoV}{field of view}

\acro{mIoU}{mean Intersection-over-Union}
\acro{ECE}{expected calibration error}

\acro{GAN}{Generative Adversarial Network}
\acro{POMCP}{Partially Observable Monte-Carlo Planning}
\acro{MCTS}{Monte-Carlo tree search}
\acro{RL}{reinforcement learning}
\acro{MRIPP}{multi-robot informative path planning}
\acro{IPP}{informative path planning}

\end{acronym}
\usepackage{multirow}
\usepackage{booktabs}
\usepackage{subcaption}
\usepackage{hyperref}
\usepackage[nameinlink, capitalize,noabbrev]{cleveref}

\DeclareMathOperator{\Tr}{Tr}

\title{\LARGE \bf
Scalable Multi-Robot Informative Path Planning for \\ Target Mapping via Deep Reinforcement Learning
}

\author{{Apoorva Vashisth$^{1}$, Manav Kulshrestha$^{1}$, Damon Conover$^{2}$,  Aniket Bera$^{1}$}\\
{\textit{$^1$Department of Computer Science, Purdue University, USA  $^2$DEVCOM Army Research Laboratory, USA}}\\
{\texttt{\{vashista, mkulshre, aniketbera\}@purdue.edu, damon.m.conover.civ@army.mil}}
}

\begin{document}

\maketitle
\thispagestyle{empty}
\pagestyle{empty}

\begin{abstract}
Autonomous robots are widely utilized for mapping and exploration tasks due to their cost-effectiveness. Multi-robot systems offer scalability and efficiency, especially in terms of the number of robots deployed in more complex environments. These tasks belong to the set of Multi-Robot Informative Path Planning (MRIPP) problems. In this paper, we propose a deep reinforcement learning approach for the MRIPP problem. We aim to maximize the number of discovered stationary targets in an unknown 3D environment while operating under resource constraints (such as path length). Here, each robot aims to maximize discovered targets, avoid unknown static obstacles, and prevent inter-robot collisions while operating under communication and resource constraints. We utilize the centralized training and decentralized execution paradigm to train a single policy neural network. A key aspect of our approach is our coordination graph that prioritizes visiting regions not yet explored by other robots. Our learned policy can be copied onto any number of robots for deployment in more complex environments not seen during training. Our approach outperforms state-of-the-art approaches by at least $\mathbf{26.2\%}$ in terms of the number of discovered targets while requiring a planning time of less than $\mathbf{2}$ sec per step. We present results for more complex environments with up to $\mathbf{64}$ robots and compare success rates against baseline planners. Our code and trained model are available at - \url{https://github.com/AccGen99/marl_ipp}. 
\end{abstract}

\section{Introduction}

Autonomous robotic systems are used in several tasks, such as search and rescue missions~\citep{berger2015innovative}, environment mapping~\citep{yoon2010cooperative}, and orchard monitoring~\citep{vashisth2024deep}. Multi-robot systems are gaining popularity in these domains due to their increased efficiency, compared to single-robot systems~\citep{cao1997cooperative} and manual approaches~\citep{su2022ai}. Key challenges for deploying multi-robot systems in these tasks include planning efficient paths for all robots to optimize the task objective, avoiding inter-robot and robot-obstacle collisions, scaling to larger multi-robot systems deployed in more complex environments, and considering communication and resource constraints.

In this work, we aim to develop a deep reinforcement learning-based, scalable, multi-robot path planning approach for discovering stationary targets in an unknown 3D environment. Here, each robot is constrained to a resource budget (e.g., battery capacity or mission time). Our considered problem setting belongs to the family of \ac{MRIPP} problems. Our 3D environment contains unknown static obstacles. Our multi-robot system consists of unmanned aerial vehicles (UAVs), where each UAV is equipped with two range-constrained modules - a unidirectional RGB-D sensor and a communication module. The challenges considered in this work include - the ability to scale to a larger number of robots deployed in more complex environments, consideration of regions explored by other robots while planning, and avoiding inter-robot and robot-obstacle collisions as the robots operate under communication constraints. Applications of our work include search and rescue missions, reconnaissance for military applications, mapping fruits in an orchard for precision agriculture, and discovering targets of interest in urban environments.

\begin{figure}[!t]
\centering
\includegraphics[width=\columnwidth]{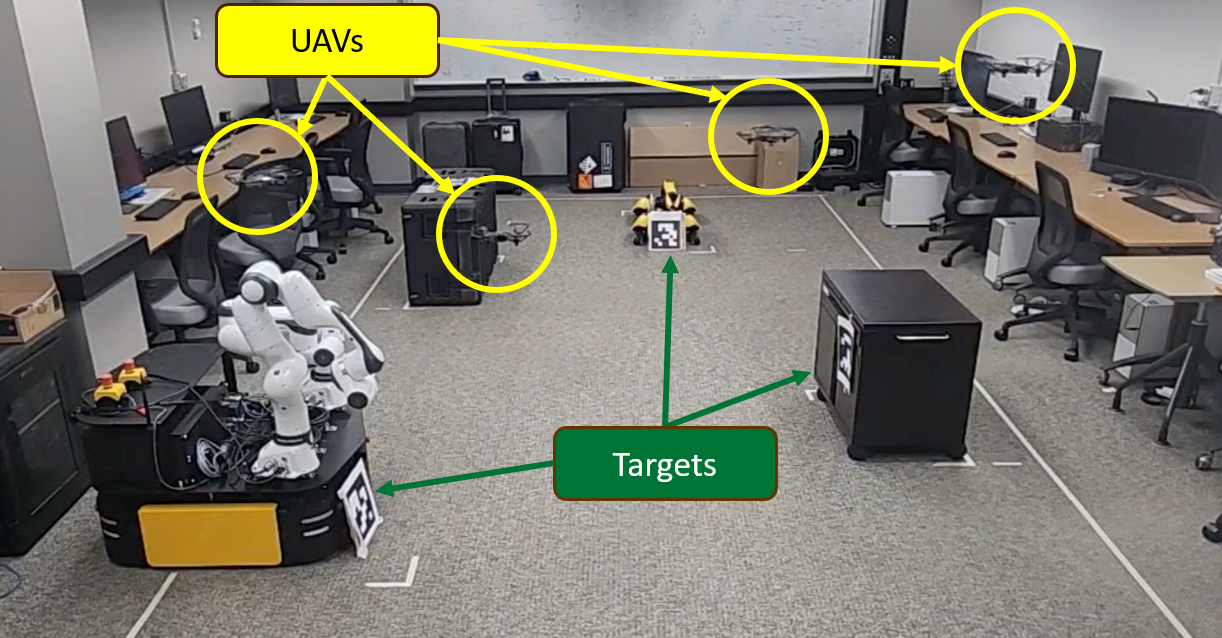}
\caption{
Our approach was implemented on Ryze Tello drones in a real-world urban monitoring scenario. We use Aruco tags as the targets to be discovered. Our approach successfully plans collision-free paths online for maximizing the number of discovered targets while under mission-time constraints. Here, $4$ Tello drones and $3$ Aruco tags are visible.}\small 
\label{F:teaser}
\end{figure}

Several approaches have been proposed for the \ac{MRIPP} problem. Classical approaches~\citep{singh2007efficient, singh2009efficient, singh2009nonmyopic, luis2024deep, hollinger2010multi} extend the single-robot planners for multi-robot systems via sequential allocation by planning path for each robot one after another, in a specific sequence. Centralized approaches~\citep{yilmaz2008path, dutta2019multi, wei2021multi, di2021multi, diop2024decoupling, barrionuevo2024informative, yoon2010cooperative, la2014cooperative, cao2013multi, cui2012pareto} plan over the joint action space of all robots. However, these planners assume availability of global communication and hence are not applicable in our problem setting with a limited communication range. Moreover, as centralized planners plan in the joint action space of all robots, they do not scale well with increasing number of robots. Recently proposed decentralized planners~\citep{viseras2019deepig, westheider2023multi, yang2023intent, yanes2023deep, best2019dec, venturini2020distributed, cui2015mutual, julian2012distributed} decouple the action space - each robot in the multi-robot system plans its own action. However, these approaches do not consider limited communication range~\citep{viseras2019deepig, yang2023intent, yanes2023deep, venturini2020distributed, cui2015mutual}, inter-robot collision avoidance~\citep{yang2023intent}, or presence of unknown static obstacles in the environment~\citep{venturini2020distributed, julian2012distributed}. To address these issues, we propose a novel decentralized approach based on deep reinforcement learning. Our approach considers the regions previously explored by other robots during planning, and constrains the planning to each robot's local region. This aids our method in not only avoiding collisions with newly discovered static obstacles, but also in preventing inter-robot collisions. As our approach is decentralized, it can be implemented on larger multi-robot systems deployed within more complex environments unseen during training.

In this work, we develop a scalable and efficient deep-reinforcement learning-based solution to the \ac{MRIPP} problem. Our approach aims to maximize the number of discovered stationary targets in an unknown 3D environment while constrained to a resource budget. We also consider avoidance of inter-robot and robot-obstacle collisions from newly discovered static obstacles. Our deep reinforcement learning based policy is decentralized in nature - a separate instance of our planner is deployed on each robot. The core aspect of our approach is the coordination graph that models the regions of the environment previously explored by other robots and constrains the robot's planning within a small local region. Our policy neural network, trained via the centralized training and decentralized execution paradigm, relies on the coordination graph for planning. As our approach is decentralized in nature, it can be implemented on a multi-robot system consisting of a large number of robots in environments not seen during training. \cref{F:teaser} illustrates our approach implemented on a multi-robot system consisting of UAVs in a real-world urban monitoring scenario.

In summary, we present the following four contributions:   
\begin{itemize}    
    \item Our coordination graph models the regions explored by other robots, enabling the policy to plan actions visiting unexplored regions of the environment.
    
    \item Our proposed reward function, aligned with the \ac{MRIPP} objective, encourages inter-robot communication.

    \item Our method enables more efficient discovery of targets ($66\%$) compared to state-of-the-art methods ($52\%$) when deployed in previously unseen environments.

    \item Our learned policy can be deployed over higher number of robots without requiring re-training.
\end{itemize}
We assess the effectiveness of our approach in an urban monitoring application in a simulator and also perform real-world experiments with Ryze Tello drones.

\section{Related Work}
Classical approaches~\citep{singh2007efficient, singh2009efficient, singh2009nonmyopic, luis2024deep, hollinger2010multi} to the \ac{MRIPP} problem attempt to extend the single-robot methods to multi-robot planning via sequential allocation~\citep{singh2007efficient}. Here, one planner plans the path for each robot one after another in an arbitrary order. These approaches decompose the environment into clusters and plan each robot's path over the clusters~\citep{singh2009efficient}, use a random sequence order of the robots for sequential allocation~\citep{hollinger2010multi}, or utilize a deep reinforcement learning approach for generating the planning order of the robots~\citep{luis2024deep}. However, these approaches assume infinite communication range and do not consider inter-robot collisions, leading to inapplicability in our problem setting.

Centralized methods~\citep{yilmaz2008path, dutta2019multi, wei2021multi, di2021multi, diop2024decoupling, barrionuevo2024informative, yoon2010cooperative, la2014cooperative, cao2013multi, cui2012pareto} introduce cooperative behavior by planning in the joint action-space of all robots. Some approaches plan paths that minimize the final uncertainty of a given uncertainty map~\citep{yilmaz2008path, dutta2019multi, wei2021multi}. Other methods decompose the environment into disjoint clusters and allocate one robot per cluster~\citep{di2021multi, yoon2010cooperative, cao2013multi, diop2024decoupling}. Other approaches rely on a consensus filter~\citep{la2014cooperative}, on deep reinforcement learning~\citep{barrionuevo2024informative}, or attempt to resolve the inter-robot collisions in pre-computed paths~\citep{cui2012pareto}. However, these approaches require availability of infinite communication range. Moreover, as they plan in the joint action space of the robots, these methods are not scalable to a large number of robots. 

Decentralized planners~\citep{viseras2019deepig, westheider2023multi, yang2023intent, yanes2023deep, best2019dec, venturini2020distributed, cui2015mutual, julian2012distributed} provide scalable solutions by allowing each robot to independently plan their next action. These approaches propose a decentralized variant of \ac{MCTS}~\citep{best2019dec}, or utilize consensus filters for encouraging cooperation among the robots~\citep{cui2015mutual, julian2012distributed}. Recently, deep reinforcement learning based decentralized planners have been developed that are not only computationally efficient at deployment but also have the capability of generalizing to similar environments not seen during training. These approaches utilize parameter sharing to encourage cooperation among robots~\citep{viseras2019deepig}, utilize the centralized training and decentralized execution paradigm~\citep{westheider2023multi, venturini2020distributed}, employ Q-Learning to learn collision avoidance behavior~\citep{yanes2023deep}, or attempt to utilize attention mechanism for modeling the paths of other robots~\citep{yang2023intent}. However, these approaches do not consider inter-robot collisions~\citep{yang2023intent, yanes2023deep} or assume an obstacle-free environment~\citep{venturini2020distributed, julian2012distributed}. A key difference of our approach with the prior works is that each robot models the regions explored by other robots within communication range and plans only within its local region. This encourages visiting unexplored regions, prevents collisions with discovered obstacles by planning in known local regions, and avoids inter-robot collisions due to planning outside of the collision range of other robots. Our results demonstrate that our approach outperforms the state-of-the-art learning and non-learning methods in our considered problem setting and is scalable to larger number of robots deployed in more complex environments not seen during training.

\section{Background}
\begin{figure*}[t]
\centering
\includegraphics[width=\linewidth]{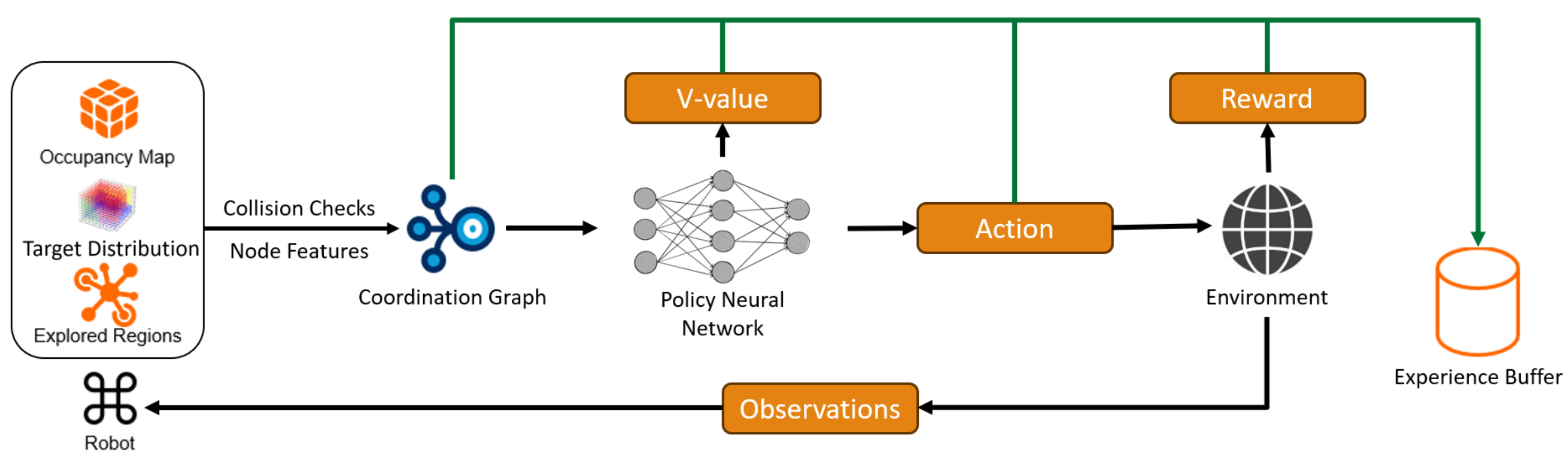}

\caption{
Overview of our deep reinforcement learning approach for the \ac{MRIPP} problem. At each time-step, our approach samples collision-free candidate actions in the robot's local region. Our coordination graph associates each candidate action with a utility value, the uncertainty of the utility value, and the exploration features modeling the regions visited by other robots. Our policy network relies on these features to output the robot's state value and the next action to execute, leading to the generation of reward and observations from the environment. Here, the black arrows indicate the robot control loop, green arrows and green boxes are the variables stored in the experience buffer for on-policy training of our policy network. 
}\small

\label{F:overview}
\end{figure*}

\subsection{Problem Setting}
In this work, we aim to maximize the number of discovered targets in a 3D environment containing undiscovered static obstacles. Our multi-robot system consists of $N \in \mathbb{Z}^+$ robots constrained to a total resource budget $B \in \mathbb{R}^+$. We model the budget as the sum of the maximum cost of the paths executed by each robot. Hence, each robot $i \in 1,\dots,N$ receives a budget of $B^i = B/N$. We model the robots as UAVs equipped with a unidirectional range sensor (e.g., RGB-D camera) and a communication module which is single-hop and range-constrained, i.e., robots communicate only when they are within distance $\rho \in \mathbb{R}^+$. To discover the targets, we assume presence of a noiseless classifier while limiting the sensing range to realistically model the reduced prediction confidence with increasing distance to target. 

\subsection{Gaussian Processes}
\label{Sec:gauss}
Gaussian processes~\citep{Rasmussen2006} are widely utilized to represent continuous distributions~\citep{yang2023intent, vashisth2024deep, cao2023catnipp} as they enable interpolation between discrete measurements. Moreover, in addition to providing predicted values, Gaussian processes have the ability to measure the uncertainty related to the predictions. These uncertainty measures are particularly valuable in our problem setting where understanding the confidence interval around a prediction is crucial for planning subsequent paths.

Given a set of $n'$ features $\mathcal{X}^* \subset \mathcal{X}$ at which a scalar value is to be inferred, a set of $n$ observed feature set $\mathcal{X'} \subset \mathcal{X}$ and the corresponding observed measurements set $\mathcal{Y}$, the mean and covariance of the GP is regressed as:
\begin{align} \notag
u &= \mu(\mathcal{X}^*) + K(\mathcal{X}^*, \mathcal{X'})[K(\mathcal{X'},\mathcal{X'})+\sigma_n^2I]^{-1}(\mathcal{Y}-\mu(\mathcal{X'}))\, \\ \notag
P &= K(\mathcal{X}^*, \mathcal{X}^*) - K(\mathcal{X}^*, \mathcal{X'})[K(\mathcal{X'},\mathcal{X'})+\sigma_n^2I]^{-1} \\ \notag & \ \ \ \ \times K(\mathcal{X}^*,\mathcal{X'})^T\,
\end{align}
where $K(\cdot)$ is a pre-trained kernel function, $\sigma^2_n$ is a hyperparameter describing the measurement noise, and $I$ is the $n\times n$ identity matrix.


\section{OUR APPROACH}
In this section, we provide details for each aspect of our proposed deep reinforcement learning based approach to the \ac{MRIPP} problem. We provide an overview of our approach in \cref{F:overview} for a robot within the multi-robot system. 

\subsection{Environment Representation}
\label{Sec:AcSp}
We define the complete action space of the robots as $\mathcal{A}$. We model the disjoint candidate action space for each robot $i$ at timestep $t$, defined as $\mathcal{A}^{i}_{t} \subset \mathcal{A}$, as a set of $j \in \{1,\dots,L\}$ actions $\mathbf{a}^{i}_{j,t} = (x^{i}_{j,t}, y^{i}_{j,t}, z^{i}_{j,t}, d^{i}_{j,t})^\top$ where $\vert \mathcal{A}^{i}_{t} \vert = L$. Here, we define the robot's 3D coordinates as $x^{i}_{j,t}, y^{i}_{j,t}, z^{i}_{j,t}\in\mathbb{R}$ and the viewing direction for the unidirectional sensor as $d^{i}_{j,t}\in\mathcal{D}$. We define a set $\mathcal{D}$ to denote possible sensor view directions. At each time-step $t$ each robot $i$ has executed an action $\mathbf{a}^i_{t-1}$. We then plan an action to execute $\mathbf{a}^i_t \in \mathcal{A}^{i}_{t}$. Similar to~\citep{vashisth2024deep}, the candidate actions are sampled randomly with a uniform distribution in the robot's $C$-neighborhood around previous pose $\mathbf{a}^{i}_{t-1}$. Here, $C$ is a constant specifying the extent of the robot's local region. To ensure inter-robot collision avoidance, we constrain the robots to not sample within collision distance $d_c \in \mathbb{R}^+$ of other robots within communication range $\rho$, and restrict that $d_c < \rho$. 

Each robot maintains an occupancy map for collision avoidance with newly discovered obstacles. We initialize the occupancy map voxels as unknown space $(1)$ and update the observed voxels as either free $(0)$ or occupied $(2)$. A voxel is occupied if it contains either a target, or a static obstacle. For robot-obstacle collision avoidance, we perform reachability checks for each candidate action along straight lines. 

Execution of an action $\mathbf{a}^{i}_{t-1}$ by robot $i$ leads to the observation of a certain number of targets at timestep $t$. To capture the relationship between an action and its corresponding number of observed targets, we define a utility function $u: \mathcal{A} \rightarrow \mathbb{R}^+$ for each candidate action. As the utility values for candidate actions are initially unknown, we utilize a Gaussian process~\citep{Rasmussen2006} to model the function $u$. The Gaussian process is trained on the utility values of executed actions and is used to regress the utility and uncertainty values of the candidate actions. To stabilize the policy learning, we normalize the observed number of targets by a constant value. The predicted uncertainty values aid our policy network in planning long-horizon paths.

At each timestep $t$, robot $i$ attempts communication with robot $j$ within communication range $\rho$. At time-step $t$, if the Euclidean distance between the robots is less than the maximum communication range $||\mathbf{a}^{i}_{t-1} - \mathbf{a}^{j}_{t-1}||_2 \leq \rho$, the robots exchange their previously visited waypoints and their current locations. Each robot maintains a Gaussian process to model the regions explored by other robots as a probability distribution over the robot's workspace. The probability and confidence values queried from the communication Gaussian process over the set of candidate actions aid our policy network in considering the regions explored by other robots while planning the next action.

\subsection{MRIPP Objective}
We model the path followed by robot $i$ as a sequence of consecutively executed actions ${\psi^i}_{0:T} = (\mathbf{a}^{i}_{0},\mathbf{a}^{i}_{1},\ldots,\mathbf{a}^{i}_{T})$ where $\mathbf{a}^{i}_{0}$ is the initial pose and $\mathbf{a}^{i}_{T}$ is the action executed upon depletion of the budget $B^i = B/N$, causing the termination of its mission. In general, the \ac{MRIPP} problem searches the space of all possible paths $\Psi^{1:N}$ for a set of optimal paths ${\psi^\ast}_{0:T} \in \Psi^{1:N}$ such that ${\psi^\ast}_{0:T} = [{\psi^1}_{0:T}, {\psi^2}_{0:T}, \ldots, {\psi^{N}}_{0:T}]$  to maximize an information-theoretic objective function:
\begin{equation}
    {\psi^\ast}_{0:T} = \mathop{{\rm {argmax}}} \limits_{{\psi^{i}}_{0:T} \in {\Psi^{1:N}}}~{\rm \sum_{i=1}^{N} \mathit{I}}({\psi^i}_{0:T}),\ {\rm {s.t.}} \,\,  c({\psi^i}_{0:T}) \leq B^i \, ,
    \label{eq:objective}
\end{equation}
where ${\rm \mathit{I}}:{\Psi} \to \mathbb{R}^+$ is the information gained upon executing the trajectory ${\psi^i}_{0:T}$ and $c~:~{\psi^i}_{0:T} \to \mathbb{R}^+$ maps the path ${\psi^i}_{0:T}$ to its execution cost.

While traversing the path ${\psi^i}_{0:t}$, the robot transitions between two consecutively executed actions over a straight line. Observations are collected at each waypoint in the path and are used to update the utility Gaussian process $(u_{util}, P_{util})$ and the occupancy map. Upon communication with a nearby robot, the communication Gaussian process $(u_{comm}, P_{comm})$ is updated with the waypoints visited by the communicating robots. Hence, due to the successive nature of the executed actions in the planned path, we model the \ac{MRIPP} problem as a sequential decision-making process. Towards the \ac{MRIPP} objective, we define a function $\zeta:\mathcal{A}\times\Psi^{1:N}\rightarrow\mathbb{R}^+$ as the number of new targets observed upon executing an action $\mathbf{a}^{i}_{t}$ by robot $i$ after following the path ${\psi^i}_{0:t-1}$. 

We define the information obtained by robot $i$ as:
\begin{equation}
    \mathit{I}({\psi^i}_{0:T}) = \sum_{t=1}^{T} \, \zeta(\mathbf{a}^{i}_{t}, {\Psi^{1:N}}_{0:t-1}) \,,
\end{equation}
%
and we aim to plan ${\psi^\ast}_{0:T}$ to maximize the information gain. 

\subsection{Reward Structure}
In order to maximize the number of discovered targets, each robot needs to balance exploration of environment with exploitation of the obtained observations. Moreover, inter-robot communication is necessary to keep track of the regions previously explored by other robots. This aids in avoiding planning of sub-optimal actions leading to re-exploration. Inspired by previous works~\citep{vashisth2024deep, tan2024ir}, we propose a new reward structure that not only considers the exploration-exploitation trade-off but also encourages inter-robot communication. Upon communication with other robots, information is exchanged about the regions previously explored by the robots. At each time-step $t$, the robot $i$ has executed the action $\mathbf{a}^{i}_{t-1}$, collected observations, communicated with the nearby robots, and receives a reward $r^{i}_{t}\in \mathbb{R}^+$. The reward function consists of an exploratory term $r^{i}_{e,t}$, an informative term $r^{i}_{u,t}$, and a communication term $r^{i}_{c,t}$ so that:

\begin{equation}
    r^{i}_{t} = \alpha r^{i}_{e,t} + \, \beta r^{i}_{u,t} + \, \gamma r^{i}_{c,t}    
\end{equation}
where:
\begin{align}
\begin{split}
    r^{i}_{e,t} &= \frac{\Tr(P_{util}^-) - \Tr(P_{util}^+)}{\Tr(P_{util}^-)} \, , \\
    r^{i}_{c,t} &= \frac{\Tr(P_{comm}^-) - \Tr(P_{comm}^+)}{\Tr(P_{comm}^-)} \, , \\
    r^{i}_{u,t} &= ~\zeta(\mathbf{a}_{t-1}, \psi_{0:t-2})
\label{E:rewardStr}
\end{split}
\end{align}
where the constants $\alpha$ and $\beta$ balance the exploration-exploitation trade-off and $\gamma$ rewards inter-robot communication. $\Tr(\cdot)$ is the matrix trace operator. Here, $P^-$ and $P^+$ indicate the prior and posterior covariance matrices of the Gaussian processes. The reduction in variance of the utility Gaussian process estimates the exploration of the environment due to the robot's own executed actions. Similarly, the variance reduction of the communication Gaussian process estimates the exploration knowledge gained from other robots. The number of new targets observed measures the information gained upon execution of action $\mathbf{a}^{i}_{t}$ by robot $i$. Scaling the reward by $\Tr(P^-)$ stabilizes the actor-critic network training~\citep{cao2023catnipp, vashisth2024deep}.

Our reward generation method ensures that each robot receives the reward reflecting the contribution of it's actions towards the global \ac{MRIPP} objective - 
\begin{itemize}
    \item Robot $i$ will not receive informative reward $r^{i}_{u,t}$ for the new targets that have been observed by another robot.

    \item At each timestep $t$ during training, we utilize a single global utility Gaussian process for all robots. The term $r^{i}_{e,t}$ is calculated for robot $i$'s action considering no action has been executed by any other robots. 

    \item The communication reward $r^{i}_{c,t}$ depends on the exploration knowledge gained by robot $i$ from other robots. Each robot has a separate instance of communication Gaussian process to generate this reward component to ensure the reward received depends on the communication performed due to execution of its own action only.
\end{itemize}

\subsection{Coordination Graph}
The \ac{MRIPP} problem considered in this work requires each robot $i$ in our multi-robot system to reason about the distribution of targets in the environment to optimize the \ac{MRIPP} objective as described in \cref{eq:objective} and the regions explored by other robots for inter-robot collision avoidance. Inspired by the dynamic graph approach for single-robot path planning~\citep{vashisth2024deep}, we propose a novel coordination graph that enables our approach to model the distribution of targets in the robot's local neighborhood, plan actions to visit regions in the environment not explored by other robots, and avoid inter-robot collisions and collisions with newly discovered static obstacles. Our policy neural network relies on the coordination graph to predict the next action to execute. 

Each robot rebuilds its coordination graph at every timestep to account for the newly obtained observations. Our coordination graph for a robot $i$ at timestep $t$ is a fully-connected graph $\mathcal{G}^{i}_{t}=(\mathcal{N}^{i}_{t}, \mathcal{E}^{i}_{t})$. The node set $\mathcal{N}^{i}_{t}$ defines the set of collision-free candidate actions. The edge set $\mathcal{E}^{i}_{t}$ defines the collision-free paths from the robot's current pose to each candidate action and each edge in the set is associated with the cost of executing the given candidate action. 

We construct the feature matrix $\mathcal{M}^{i}_{t}$ for robot $i$ corresponding to its coordination graph as the input to our policy neural network. The features consist of the candidate actions, the utility and uncertainty values of candidate actions regressed from the Gaussian process modeling the utility, and the probability and uncertainty values queried from the communication Gaussian process. The $n^{\mathrm{th}}$ row of $\mathcal{M}^{i}_{t}$ relates to the $n^{\mathrm{th}}$ candidate action of robot $i$ at timestep $t$:
\begin{align}
\begin{split}
    \mathbf{M}^{i}_{t}(n)= &[\mathbf{a}^{i}_{n,t},  \, u_{utility}(\mathbf{a}^{i}_{n,t}),  \, P_{utility}(\mathbf{a}^{i}_{n,t}, \mathbf{a}^{i}_{n,t}), \, \\ &u_{comm}(\mathbf{a}^{i}_{n,t}, t),  \,
    P_{comm}(\mathbf{a}^{i}_{n,t}, \mathbf{a}^{i}_{n,t})] \, , \notag
\label{E:augGr}
\end{split}
\end{align}
where $\mathbf{a}^{i}_{n,t} = [x^{i}_{n,t}, \,  y^{i}_{n,t},  \, z^{i}_{n,t},  \, d^{i}_{n,t}]^\top$, $u_{utility}(\mathbf{a}^{i}_{n,t})$ and $P_{utility}(\mathbf{a}^{i}_{n,t}, \mathbf{a}^{i}_{n,t})$ are the regressed utility and uncertainty values for candidate action $\mathbf{a}^{i}_{n,t}$, and $u_{comm}(\mathbf{a}^{i}_{n,t})$ and $P_{comm}([\mathbf{a}^{i}_{n,t}], [\mathbf{a}^{i}_{n,t}])$ are the regressed probability and confidence values from the communication Gaussian process modeling the locations of other robots at timestep $t$.

\subsection{Policy Neural Network}
\label{SS:rl_network}
Our coordination graph models each robot's collision free action space and aids the deep reinforcement learning policy in reasoning about the robot's current knowledge of the environment. 
As the utility Gaussian process can only model the greedy action selection through utility regression, deep reinforcement learning is essential for achieving the balance in short-term exploitation of obtained information and the long-term exploration of the unknown environment.

At each timestep $t$, each robot $i$ in our multi-robot system utilizes an attention-based neural network~\citep{cao2023catnipp} to model the planning policy~$\pi(\mathcal{G}^{i}_{t}, {\psi^i}_{0:t-1}, \tilde{B}^{i})$. The planning policy outputs the probability distribution over the candidate actions in the robot's candidate action set $\mathcal{A}^{i}_{t}$. The policy relies on the feature matrix $\mathbf{M}^{i}_{t}$ of the current coordination graph $\mathcal{G}^{i}_{t}$, path executed so far ${\psi^i}_{0:t-1}$ and the remaining budget $\tilde{B}^i$. The network structure consists of an encoder and a decoder module. The encoder models the information distribution obtained from observations and the environment explored so far by learning the dependencies among the candidate actions in $\mathcal{G}^{i}_{t}$, forming the context over collected observations. The decoder utilizes the learned context features from the encoder, the planning state, and the budget mask to output the probability distribution over the set of candidate actions $\mathcal{A}^{i}_{t}$. The planning state consists of the path executed by the robot so far ${\psi^i}_{0:t-1}$ and the remaining budget $\tilde{B}^i$. The budget mask aids in filtering out candidate actions leading to the violation of the budget constraint. Additionally, the decoder module estimates the value function of the current state. The estimated value, executed actions, coordination graphs, planning states, and rewards generated by actions of the robots throughout the training episode are collected in the experience buffer for on-policy actor-critic reinforcement learning under centralized training and decentralized execution paradigm. In this work, we use proximal policy optimization~\citep{schulman2017proximal} due to its stability and sample efficiency. During deployment, at each time step and for each robot, we execute the most informative action. 


\section{RESULTS}

\subsection{Setup}
\label{Sec:implement}
\textbf{Environment}. We test our approach in an urban monitoring scenario consisting of buildings and windows. We represent the environment internally as bounded within a scale-agnostic unit cube. Each robot in our multi-robot system maintains an occupancy map for collision avoidance with static obstacles. We initialize the occupancy map as unknown space and update the free space or occupied space based on obtained observations. Our training environment consists of regularly spaced buildings with the windows generated randomly on the buildings. However, our test environments consist of buildings generated at random locations.

\textbf{Hyperparameters}. We tune the hyperparameters of our Gaussian processes in a small representative environment. We use the Matérn~$1/2$ kernel function for the Gaussian processes. For the reward structure defined in ~\Cref{eq:objective}, we choose $\alpha=20.0$ and $\delta=0.02$ so that both the exploratory $r^{i}_{e,t}$ and utility reward $r^{i}_{u,t}$ terms lie numerically in the range $[0,1]$. In order to promote \ac{MRIPP} objective over the inter-robot communication, we use $\gamma = 1.0$ to keep the numerical value of $r^{i}_{c,t}$ lower than the other terms.

\textbf{Robot Configuration}. We consider each robot as a UAV platform equipped with an RGB-D camera having $90^{\circ}$ field of view. We model the reduction in the confidence of target identification with increasing distance by limiting the camera sensing range to $24\%$ of the environment size. The UAVs can communicate at a maximum communication distance $\rho=0.3$. The sensor viewpoint set $\mathcal{D}$ is discretized as $\{0, \frac{\pi}{2}, \pi, \frac{3\pi}{2}\}$ radians. However, our approach supports extension to finer discretizations by extending the set $\mathcal{D}$.

\textbf{Training}. We generate multiple training episodes to populate our experience buffer. Each training episode consists of a multi-UAV system with a total budget $B$. Our policy is trained in a structured environment and then transferred to a randomized environment for testing. While we fix the number of buildings during training, the number of windows is varied in $[200, 250]$. The start action for each robot is $\mathbf{a}_0 = (0.0, 0.0, 0.0, \frac{\pi}{2})$. We set $L=80$ for each robot's coordination graph. Since we normalize the internal environment representation, our budget value $B$ is unitless. For each training episode, $B$ is a randomly generated real value in the range $[7.0, 9.0]$. Each episode is constrained to a maximum of $256$ timesteps. To speed-up the training process, we run $36$ parallel environment instances and train our policy network over $8$ epochs with a batch size of $1024$. We utilize Adam optimizer with a learning rate of $10^{-4}$, decaying by a factor of $0.96$ after every $512$ optimization steps. The policy gradient epsilon-clip parameter is set to $0.2$. We train our policy network on a computing cluster equipped with Intel(R) Xeon(R) CPUs @ 3.60GHz and one NVIDIA A30 Tensor Core GPU. We require $\sim 120,000$ environment interactions for our policy to converge.

\subsection{Baseline Comparison}
\label{Sec:BaseComp}
In this section, we compare the performance of our approach against state-of-the-art learning and non-learning baselines. We utilize fixed random seeds to generate $25$ test environments. We enable global communications for the baselines and present the performance of our approach with both global communication $(\rho=\infty)$ and restricted communication $(\rho=0.1)$. We deploy $N=3$ UAVs with a total budget of $B=10.00$ and run $10$ trials corresponding to each randomly generated environment, leading to a total of $250$ tests. Our baselines include: (i)~Intent with destination modeling~\citep{yang2023intent} as a zero-shot greedy policy (Intent dest.), (ii)~CAtNIPP~\citep{cao2023catnipp} with a zero-shot policy (CAtNIPP g.), (iii)~non-learning Monte Carlo Tree Search~\citep{ott2023sequential} (MCTS), (iv)~non-learning Rapidly exploring random Information Gathering Trees~\citep{hollinger2014sampling} (RIG-Tree), and (v)~a random policy (random agent). As MCTS, and RIG-Tree are single robot planners, we extend them to multi-robot planning via sequential allocation~\citep{singh2007efficient}. We consider the metric of percentage of targets discovered, as well as provide the average planning time per step. Additionally, as modification of these approaches to account for collision-free path planning is non-trivial, we allow the UAVs for each planner to ignore obstacle-UAV and inter-robot collisions, ensuring a fair comparison within this set of experiments.

\begin{figure}[t]
\centering
\includegraphics[width=0.9\linewidth]{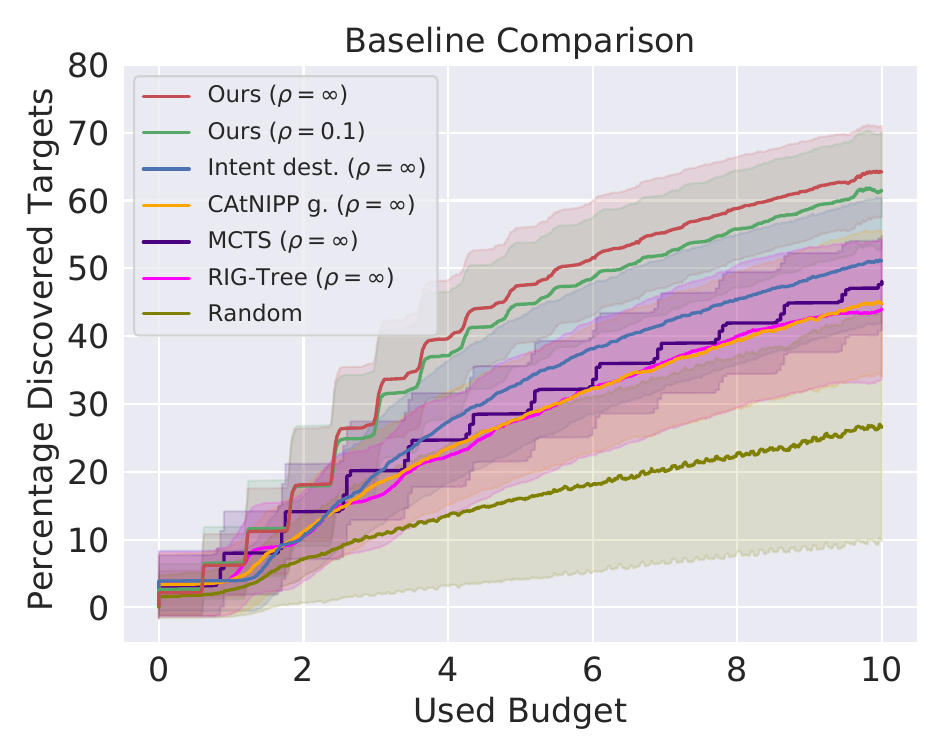}
\caption{
Comparison of our approach with other baselines in an urban environment. Our performance metric is the percentage of targets discovered during the episode. The solid lines represent the mean values across $250$ trials, while the shaded areas denote the standard deviations.
}\small
\label{F:baseComp}
\end{figure}

\begin{table}[!t]
    \vspace{3mm}
    \small
    \centering
    \caption{Results of our approach compared with other learning and non-learning baselines in an urban monitoring scenario.}
\begin{tabular}{l|p{1.75cm}<{\centering}|p{1cm}<{\centering}p{1cm}<{\centering}}
         \multicolumn{1}{c|}{\multirow{1}{*}{Baseline}} & \multicolumn{1}{c|}{$\%$ targets} & \multicolumn{1}{c}{Time (s)} \\
         \midrule
         Our approach $(\rho=\infty)$ & $\mathbf{66.00 \pm 5.59}$ & $\mathbf{1.58}$\\ 
         Our approach $(\rho=0.1)$ & $\mathbf{62.16 \pm 8.25}$ & $\mathbf{1.58}$\\ 
         Intent dest. $(\rho=\infty)$ & $52.31 \pm 9.29$ & $25.39$\\ 
         CAtNIPP g. $(\rho=\infty)$ & $45.31 \pm 10.59$ & $3.84$\\ 
         MCTS $(\rho=\infty)$ & $48.50 \pm 7.81$ & $153.49$ \\
         RIG-Tree $(\rho=\infty)$ & $47.38 \pm 10.43$ & $60.46$ \\
         Random agent & $31.83 \pm 15.23$ & $0.06$\\ 
    \end{tabular}
    \label{T:baseTable}
\end{table}

Our results shown in \cref{F:baseComp} and \cref{T:baseTable} indicate that our method outperforms the considered baselines in terms of the number of discovered targets. Note that introduction of communication constraint leads to a drop in the performance of our approach, however we still outperform the considered baselines operating with global communications. This could be attributed to our coordination graph explicitly representing the regions explored by other robots, leading to planning actions visiting unexplored regions that provide performance improvement in our approach over the baselines that do not model the regions visited by other robots. Note that deep reinforcement learning based methods are significantly more time-efficient than non-learning approaches, justifying their use over non-learning methods for real-time applications.

\subsection{Ablation Studies}
We study the impact of our communication Gaussian process and our new reward structure on the performance of our approach via the metric of percentage discovered targets.

\begin{table}[t]
    \vfill
    \small
    \centering
    \caption{Ablation study for modeling of explored regions.}

\begin{tabular}{l|p{1.75cm}<{\centering}p{1.5cm}<{\centering}}
        \multicolumn{1}{c|}{\multirow{1}{*}{Approach}} & \multicolumn{1}{c}{$\%$ targets} \\
         \midrule
         With modeling explored regions & $66.00 \pm 5.59$\\
         Without modeling explored regions & $61.97 \pm 7.34$\\ 
    \end{tabular}
    \label{T:ablation_str}
\vspace{-5mm}
\end{table}

\begin{table}[t]
    \vspace{3mm}
    \small
    \centering
    \caption{Ablation study for reward structure.}

\begin{tabular}{l|p{1.75cm}<{\centering}p{1.5cm}<{\centering}}
        \multicolumn{1}{c|}{\multirow{1}{*}{Approach}} & \multicolumn{1}{c}{$\%$ targets} \\
         \midrule
         With communication reward & $66.00 \pm 5.59$\\
         Without communication reward & $62.92 \pm 7.01$\\
    \end{tabular}
    \label{T:ablation_rew}
\end{table}

\textbf{Communication Gaussian Process}. To evaluate the impact of modeling the unexplored regions of the environment on the performance of our approach, we compare our approach trained with and without the communication Gaussian process. Our results in \cref{T:ablation_str} show that the performance improves when the communication Gaussian process is included. Furthermore, an unpaired t-test conducted between the two groups ($n = 250$ each) yielded a p-value of $3.91 \times 10^{-9}$, indicating a statistically significant difference well below the conventional threshold of $0.05$. Hence, we conclude that our policy neural network learns to reason about the unexplored regions during planning.

\textbf{Communication Reward}. To evaluate the impact of the communication reward term $r^{i}_{c,t}$ in \cref{E:rewardStr} on the performance of our approach, we compare the performance of our policies trained with $\gamma = 0.0$ and $\gamma = 1.0$. Our results in \cref{T:ablation_rew} show that the performance improves upon inclusion of the communication reward. Again, the difference between the two groups of $250$ tests each is statistically significant, with a p-value of $5.11\times10^{-6}$, significantly less than $0.05$, providing strong evidence against the null hypothesis. Hence, our new reward structure promotes inter-robot communication and leads to improved performance. 

\subsection{Scalability}
We compare the ability of our approach to scale to larger environments and more number of robots $N$ in the multi-robot system with other approaches. Our policy learned in the small training environment with $N=3$ robots is evaluated in larger environments and varying number of robots not seen during training. We present results for test environments that are approximately $3\times,8\times,$ and $16\times$ larger than the training environment. We consider $N \in (16,\ 32,\ 48,\ 64)$ and conduct $100$ tests in each environment for every $N$.

We compare the performance of our approach, deployed with communication distance $\rho \in(0.2, 0.15, 0.12)$ in $3\times, 8\times, 16\times$ environment respectively, with (i)~CAtNIPP~\citep{cao2023catnipp} with global observability and (ii)~a random policy. We train CAtNIPP in the training environment described in~\cref{Sec:implement}. We do not consider the intent baseline due to its compute intensive nature for large multi-robot systems. We do not evaluate the non-learning methods, as the sequential allocation based time-intensive nature leads to infeasible computation times for large multi-robot systems.

\begin{figure}[t]
\centering
\includegraphics[width=0.8\linewidth]{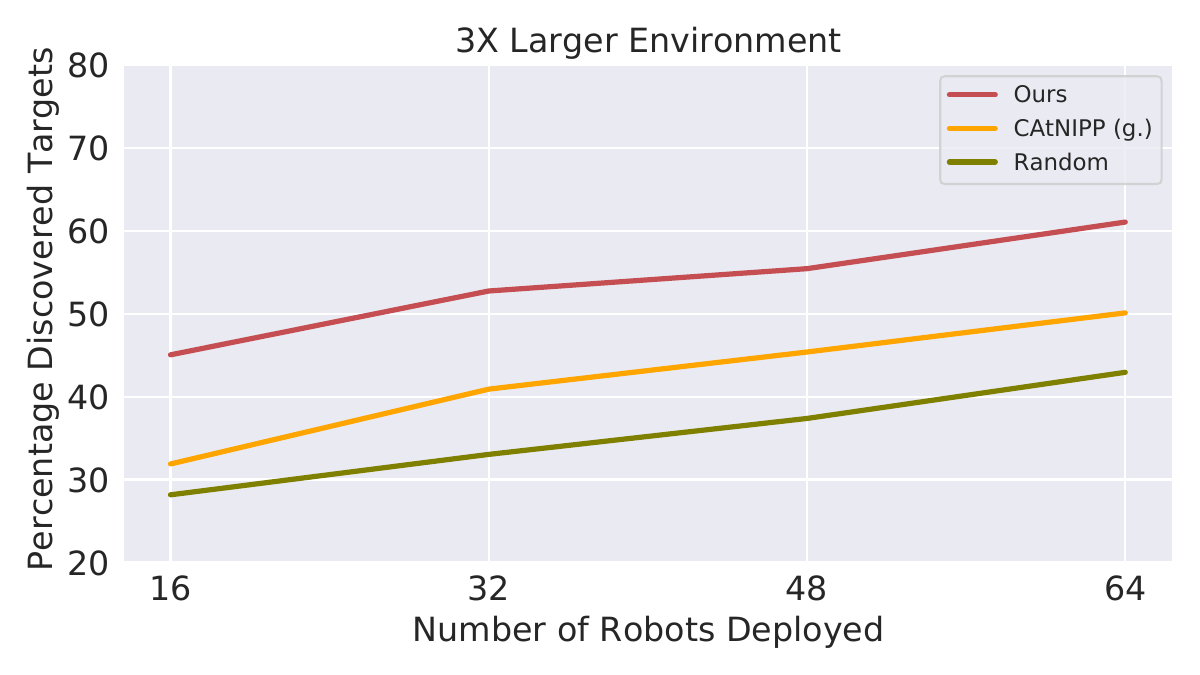}
\includegraphics[width=0.8\linewidth]{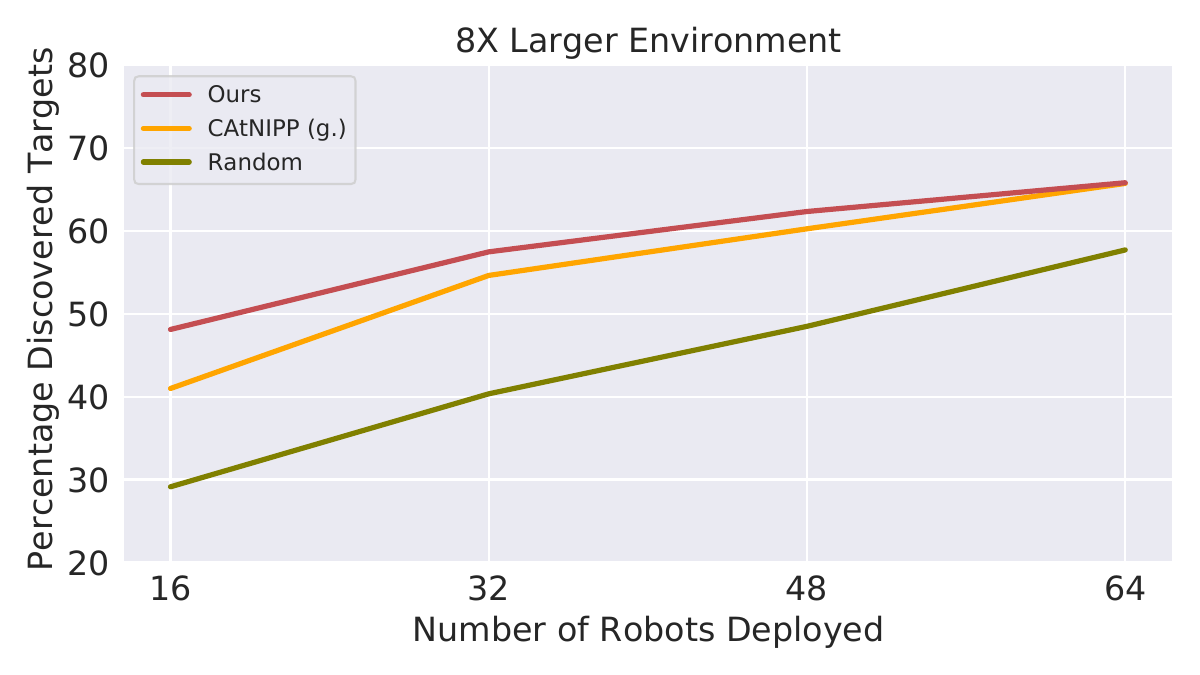}
\includegraphics[width=0.8\linewidth]{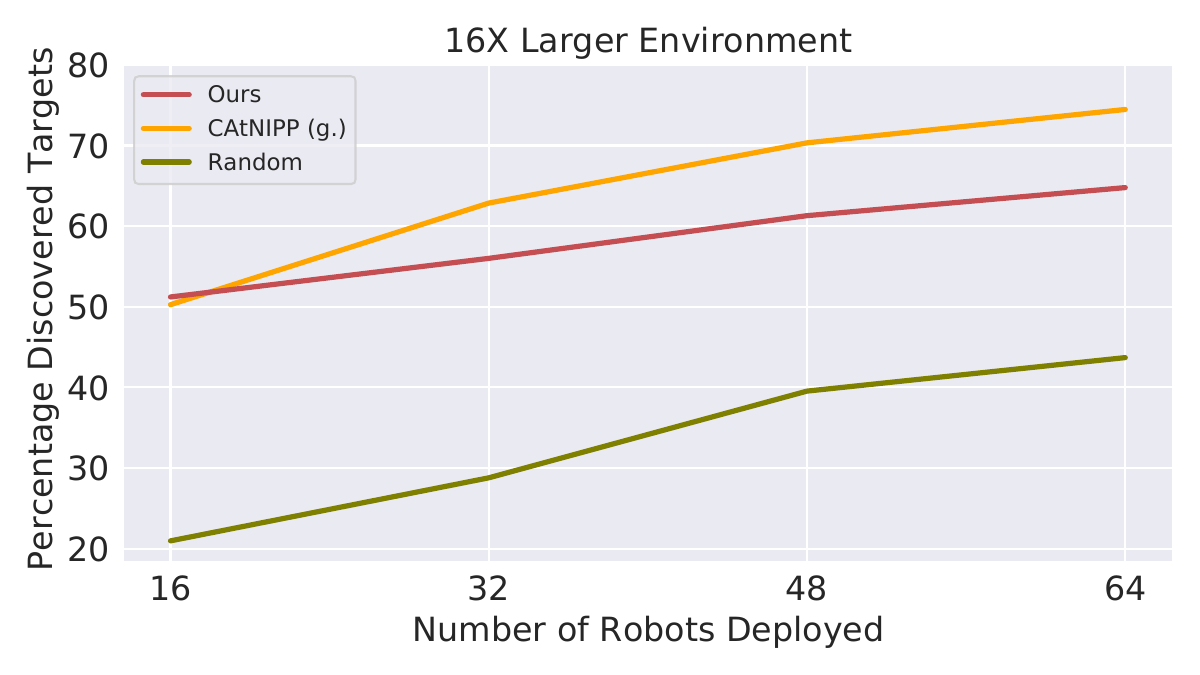}
\caption{
Our approach outperforms other baselines in terms of percentage discovered targets in a $3\times,\ 8\times,\ $and $16\times$ larger environments. The x-axis indicates the number of robots in the multi-robot system during the test. The solid lines represent the mean values across $100$ trials.}\small
\label{F:scale_all}
\end{figure}

\cref{F:scale_all} demonstrates the results for this set of experiments. Our approach strongly outperforms the considered baselines in $3\times$ environment, slightly outperforms in $8\times$ environment, and is outperformed in $16\times$ environment. The success of our planner in $3\times$ and $8\times$ environments can be attributed to the modeling of unexplored regions, leading to planning of more informative paths by robots in our multi-robot system as compared to other approaches. However, in the $16\times$ environment, the instances of communication with robots located further away drastically reduces, causing our approach to be outperformed by the CAtNIPP planner with global observability. Future work will explore more complex communication paradigms to mitigate this issue. 

\begin{table}[t]
    \vfill
    \small
    \centering
    \caption{Comparison of our deep reinforcement learning-based approach against baselines in an urban environment.}

\begin{tabular}{l|p{1.75cm}<{\centering}p{1.5cm}<{\centering}}
        \multicolumn{1}{c|}{\multirow{1}{*}{Approach}} & \multicolumn{1}{c}{$\%$ targets} \\
         \midrule
         Our Approach & $66.46 \pm 11.08$\\
         Random Planner & $28.94 \pm 8.70$\\ 
    \end{tabular}
    \label{T:ue4Comp}

\vspace{-3mm}
\end{table}

\subsection{Simulation}
\begin{figure}[!t]
\centering
\includegraphics[width=0.85\linewidth]{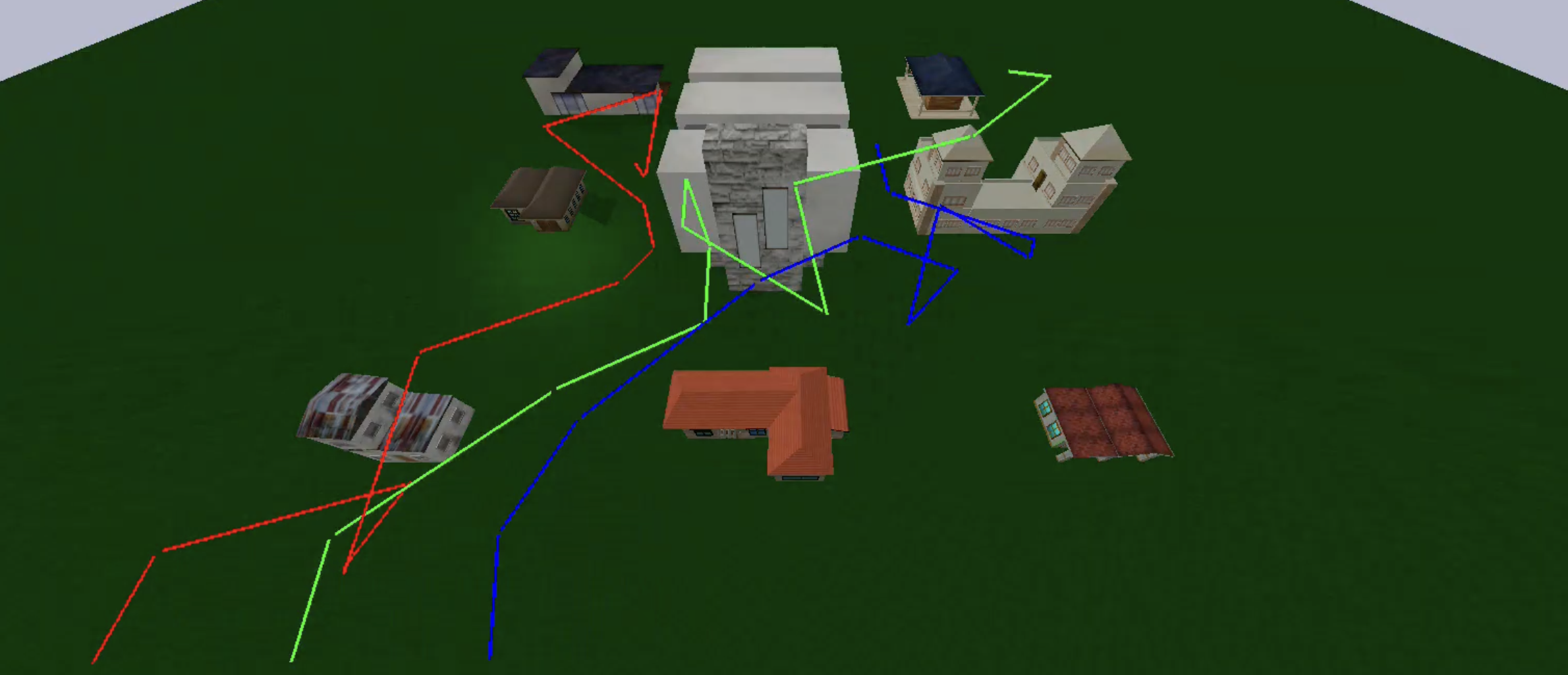}
\caption{Our approach implemented in an urban simulation environment. We place $3$ UAVs and $161$ targets in the environment and trace each UAV's path with colored tracelines.}\small
\label{F:sims}
\vspace{-3mm}
\end{figure}

We demonstrate the applicability of our deep reinforcement learning approach in an urban monitoring scenario. We use the gym-PyBullet-drones~\citep{panerati2021learning} simulator to accurately model UAV physics. Our simulation environment is built using the Houses3K dataset~\citep{peralta2020next} and is bounded by a $60\,\text{m}\times60\,\text{m}\times30\,\text{m}$ cuboid as shown in~\cref{F:sims}. We assume perfect localization and use ground truth target discovery. The $3$ UAVs move at a maximum speed of $1$\,m/s.  

We compare the performance of our approach with a random planner that reflects the performance lower bound. We do not implement other baselines considered in~\cref{Sec:BaseComp} as modifying these approaches for avoiding inter-robot collisions and consideration of the presence of unknown obstacles in the environment is a non-trivial task. Here, our evaluation metric is the percentage of windows discovered by the robots. To ensure every discovered target is counted only once, we record the coordinates of discovered targets. Our results are reported for missions with a budget of $7.0$ units in \cref{T:ue4Comp}. Our approach outperforms the random planner.

\subsection{Implementation}
We demonstrate the real-world applicability of our method on a multi-robot system for target discovery as illustrated in \cref{F:teaser}. We carried out experiments on $4$ Ryze Tello drones in a $7.62 \times 3.25 \times 2.4 \ \text{m}^3$ arena containing randomly placed obstacles and $6$ Aruco tags as targets. 


\section{Conclusion}
We present a novel deep reinforcement learning approach for the \ac{MRIPP} problem in an unknown 3D environment. Our coordination graph-based approach models the unexplored regions of the environment for efficient target discovery. We present experimental results to support that: (i)~our coordination graph encourages exploration of unknown regions of the environment, (ii)~our reward function encourages inter-robot communication, (iii)~our approach outperforms the state-of-the-art baselines in environments unseen during training, and (iv)~our learned policy scales to larger multi-robot systems and more complex environments. We evaluate the performance of our approach in a UAV-based urban mapping scenario in a simulator, as well as conduct real robot experiments to demonstrate the practical applicability. Future research directions include extension to multi-robot pathfinding, task allocation, and cooperative communication.










\bibliographystyle{IEEEtranN}
\footnotesize
\bibliography{root}

\end{document}